\documentclass[10pt,twocolumn,letterpaper]{article}

\usepackage{iccv}
\usepackage{times}
\usepackage{graphicx}
\usepackage{amsmath}
\usepackage{amssymb}
\usepackage{bm}
\usepackage{tikz,cite}
\usetikzlibrary{shapes.geometric, arrows}

\tikzstyle{node} = [rectangle, rounded corners, minimum width=1cm, minimum height=1cm,text centered, draw=black, fill=red!20]
\tikzstyle{arrow} = [thick,->,>=stealth]

\newcommand{\logs}{\textrm{logs}}
\newcommand{\barr}{\mathbf{\bar r}}
\newcommand{\R}{\mathbb{R}}

\usepackage[pagebackref=true,breaklinks=true,colorlinks,bookmarks=false]{hyperref}

\iccvfinalcopy 

\def\iccvPaperID{15} 

\begin{document}

\title{3D Morphable Models as Spatial Transformer Networks}

\author{Anil Bas\textsuperscript{*}, Patrik Huber\textsuperscript{\textdagger}, William A. P. Smith\textsuperscript{*}, Muhammad Awais\textsuperscript{\textdagger}, Josef Kittler\textsuperscript{\textdagger}\\
\textsuperscript{*}Department of Computer Science, University of York, UK\\
\textsuperscript{\textdagger}Centre for Vision, Speech and Signal Processing, University of Surrey, UK\\
{\tt\small \{ab1792,william.smith\}@york.ac.uk, \{p.huber,m.a.rana,j.kittler\}@surrey.ac.uk}
}

\maketitle
\thispagestyle{empty}

\begin{abstract}
In this paper, we show how a 3D Morphable Model (i.e. a statistical model of the 3D shape of a class of objects such as faces) can be used to spatially transform input data as a module (a 3DMM-STN) within a convolutional neural network. This is an extension of the original spatial transformer network in that we are able to interpret and normalise 3D pose changes and self-occlusions. The trained localisation part of the network is independently useful since it learns to fit a 3D morphable model to a single image. We show that the localiser can be trained using only simple geometric loss functions on a relatively small dataset yet is able to perform robust normalisation on highly uncontrolled images including occlusion, self-occlusion and large pose changes.
\end{abstract}

\section{Introduction}\label{sec:intro}

Convolutional neural networks (CNNs) are usually trained with such large amounts of data that they can learn invariance to scale, translation, in-plane rotation and, to a certain degree, out-of-plane rotations, without using any explicit geometric transformation model. However, most networks do require a rough bounding box estimate as input and don't work for larger variations. Recently, Jaderberg \etal \cite{DBLP:conf/nips/JaderbergSZK15} proposed the Spatial Transformer Network (STN) - a module that can be incorporated into a neural network architecture, giving the network the ability to explicitly account for the effects of pose and nonrigid deformations (which we refer to simply as ``pose'').
An STN explicitly estimates pose and then resamples a specific part of the input image to a fixed-size output image. It is thus able to work on inputs with larger translation and pose variation in general, since it can explicitly compensate for it, and feed a transformed region of interest to the subsequent neural network layers. By exploiting and ``hard-coding'' knowledge of geometric transformation, the amount of training data and the required complexity of the network can be vastly reduced.

In this paper, we show how to use a 3D morphable model as a spatial transformer network (we refer to this as a 3DMM-STN). In this setting, the locations in the input image that are resampled are determined by the 2D projection of a 3D deformable mesh. Hence, our 3DMM-STN estimates both 3D shape and pose. This allows us to explicitly estimate and account for 3D rotations as well as self occlusions. The output of our 3DMM-STN is a resampled image in a flattened 2D texture space in which the images are in dense, pixel-wise correspondence. Hence, this output can be fed to subsequent CNN layers for further processing. We focus on face images and use a 3D morphable face model \cite{blanz1999morphable,DBLP:conf/avss/PaysanKARV09}, though our idea is general and could be applied to any object for which a statistical 3D shape model is available (though note that the loss functions proposed in Sections \ref{sec:bilateral} and \ref{sec:siamese} do assume that the object is bilaterally symmetric).
We release source code for our 3DMM-STN in the form of new layers for the MatConvNet toolbox~\cite{DBLP:conf/mm/VedaldiL15}\footnote{The source code is available at \url{https://github.com/anilbas/3DMMasSTN}.}.

\subsection{Related work} \label{sec:related-work}

In a lot of applications, the process of pose normalisation and object recognition are disjoint. For example, in the breakthrough deep learning face recognition paper DeepFace, Taigman \etal \cite{DBLP:conf/cvpr/TaigmanYRW14} use a 3D mean face as preprocessing, before feeding the pose-normalised image to a CNN.

\paragraph{Spatial transformers} The original STN \cite{DBLP:conf/nips/JaderbergSZK15} aimed to combine these two processes into a single network that is trainable end to end. The localiser network estimated a 2D affine transformation that was applied to the regular output grid meaning the network could only learn a fairly restricted space of transformations. Jaderberg \etal \cite{DBLP:conf/nips/JaderbergSZK15} also proposed the concept of a 3D transformer, which takes 3D voxel data as input, applies 3D rotation and translation, and outputs a 2D projection of the transformed data. Working with 3D (volumetric data) removes the need to model occlusion or camera projection parameters. In contrast, we work with regular 2D input and output images but transform them via a 3D model.

A number of subsequent works were inspired by the original STN.
Yan \etal \cite{yan2016perspective} use an encoder-decoder architecture in which the encoder estimates a 3D volumetric shape from an image and is trained by combining with a decoder which uses a perspective transformer network to compute a 2D silhouette loss.
Handa \etal \cite{Handa:etal:ECCVW16} present the gvnn (Geometric Vision with Neural Networks) toolbox that, like in this paper, has layers that explicitly implement 3D geometric transformations. However, their goal is very different to ours. Rather than learning to fit a statistical shape model, they seek to use 3D transformations in low level vision tasks such as relative pose estimation. Chen \etal \cite{DBLP:conf/eccv/ChenHW016} use a spatial transformer that applies a 2D similarity transform as part of an end to end network for face detection. Henriques and Vedaldi \cite{DBLP:journals/corr/HenriquesV16} apply a spatial warp prior to convolutions such that the convolution result is invariant to a class of two-parameter spatial transformations. Like us, Yu \etal \cite{DBLP:conf/eccv/YuZC16} incorporate a parametric shape model, though their basis is 2D (and trainable), models only sparse shape and combines pose and shape into a single basis. They use a second network to locally refine position estimates and train end to end to perform landmark localisation.
Bhagavatula \etal \cite{bhagavatula2017faster} fit a generic 3D face model and estimate face landmarks, before warping the projected face model to better fit the landmarks. They estimate 2D landmarks in a 3D-aware fashion, though they require known landmarks for training.

\paragraph{Analysis-by-synthesis} Our localiser learns to fit a 3DMM to a single image.
This task has traditionally been posed as a problem of analysis-by-synthesis and solved by optimisation. The original method \cite{blanz1999morphable} used stochastic gradient descent to minimise an appearance error, regularised by statistical priors. Subsequent work used a more complex feature-based objective function \cite{romdhani2005estimating} and the state-of-the-art method uses Markov Chain Monte Carlo for probabilistic image interpretation \cite{schonborn2017markov}.

\paragraph{Supervised CNN regression}
Analysis-by-synthesis approaches are computationally expensive, prone to convergence on local minima and fragile when applied to in-the-wild images. For this reason, there has been considerable recent interest in using CNNs to directly regress 3DMM parameters from images. The majority of such work is based on supervised learning. Jourabloo and Liu \cite{cvpr16/jourabloo} fit a 3DMM to detected landmarks and then train a CNN to directly regress the fitted pose and shape parameters. Tr\~{a}n \etal \cite{tran2017regressing} use a recent multi-image 3DMM fitting algorithm \cite{piotraschke2016automated} to obtain pooled 3DMM shape and texture parameters (i.e. the same parameters for all images of the same subject). They then train a CNN to directly regress these parameters from a single image. They do not estimate pose and hence do not compute an explicit correspondence between the model and image. Kim \etal \cite{kim2017inversefacenet} go further by also regressing illumination parameters (effectively performing inverse rendering) though they train on synthetic, rendered images (using a breeding process to increase diversity). They estimate a 3D rotation but rely on precisely cropped input images such that scale and translation is implicit. Richardson \etal \cite{richardson20163d} also train on synthetic data though they use an iteratively applied network architecture and a shape-from-shading refinement step to improve the geometry. Jackson \etal \cite{jackson2017large} regress shape directly using a volumetric representation.

The DenseReg \cite{guler2017densereg} approach uses fully convolutional networks to directly compute dense correspondence between a 3D model and a 2D image. The network does not explicitly estimate or model 3D pose or shape (though these are implied by the correspondence) and is trained by using manually annotated 2D landmarks to warp a 3D template onto the training images (providing the supervision). 
Sela \etal \cite{sela2017unrestricted} also use a fully convolutional network to predict correspondence and also depth. They then merge the model-based and data-driven geometries for improved quality.

The weakness of all of these supervised approaches is that they require labelled training data (i.e. images with fitted morphable model parameters). If the images are real world images then the parameters must come from an existing fitting algorithm in which case the best the CNN can do is learn to replicate the performance of an existing algorithm. If the images are synthetic with known ground truth parameters then the performance of the CNN on real world input is limited by the realism and variability present in the synthetic images. Alternatively, we must rely on 3D supervision provided by multiview or RGBD images, in which case the available training data is vastly reduced.

\paragraph{Unsupervised CNN regression}
Richardson \etal \cite{richardson2017learning} take a step towards removing the need for labels by presenting a semi-supervised approach. They still rely on supervised training for learning 3DMM parameter regression but then refine the coarse 3DMM geometry using a second network that is trained in an unsupervised manner. Very recently, Tewari \etal \cite{tewari2017mofa} presented MoFA, a completely unsupervised approach for training a CNN to regress 3DMM parameters, pose and illumination using an autoencoder architecture. The regression is done by the encoder CNN. The decoder then uses a hand-crafted differentiable renderer to synthesise an image. The unsupervised loss is the error between the rendered image and the input, with convergence aided by losses for priors and landmarks. Note that the decoder is exactly equivalent to the differentiable cost function used in classical analysis-by-synthesis approaches. Presumably, the issues caused by the non-convexity of this cost function are reduced in a CNN setting since the gradient is averaged over many images.

While the ability of \cite{tewari2017mofa} to learn from unlabelled data is impressive, there are a number of limitations. The complexity required to enable the hand-crafted decoder to produce photorealistic images of any face under arbitrary real world illumination, captured by a camera with arbitrary geometric and photometric properties, is huge. Arguably, this has not yet been achieved in computer graphics. Moreover, the 3DMM texture should only capture intrinsic appearance parameters such as diffuse and specular albedo (or even spectral quantities to ensure independence from the camera and lighting). Such a model is not currently available.

\subsection{Contributions}
In this paper we propose a purely geometric approach in which only the shape component of a 3DMM is used to geometrically normalise an image. Unlike \cite{tran2017regressing,kim2017inversefacenet,richardson20163d,guler2017densereg,sela2017unrestricted,jackson2017large,cvpr16/jourabloo}, our method can be trained in an unsupervised fashion, and thus does not depend on synthetic training data or the fitting results of an existing algorithm. In contrast to \cite{tewari2017mofa}, we avoid the complexity and potential fragility of having to model illumination and reflectance parameters. Moreover, our 3DMM-STN can form part of a larger network that performs a face processing task and is trained end to end. Finally, in contrast to all previous 3DMM fitting networks, the output of our 3DMM-STN is a 2D resampling of the original image which contains all of the high frequency, discriminating detail in a face rather than a model-based reconstruction which only captures the gross, low frequency aspects of appearance that can be explained by a 3DMM.


\section{3DMM-STN}

\begin{figure}[t]
\begin{center}
\includegraphics[clip, trim=152px 192px 150px 190px, width=1.00\columnwidth]{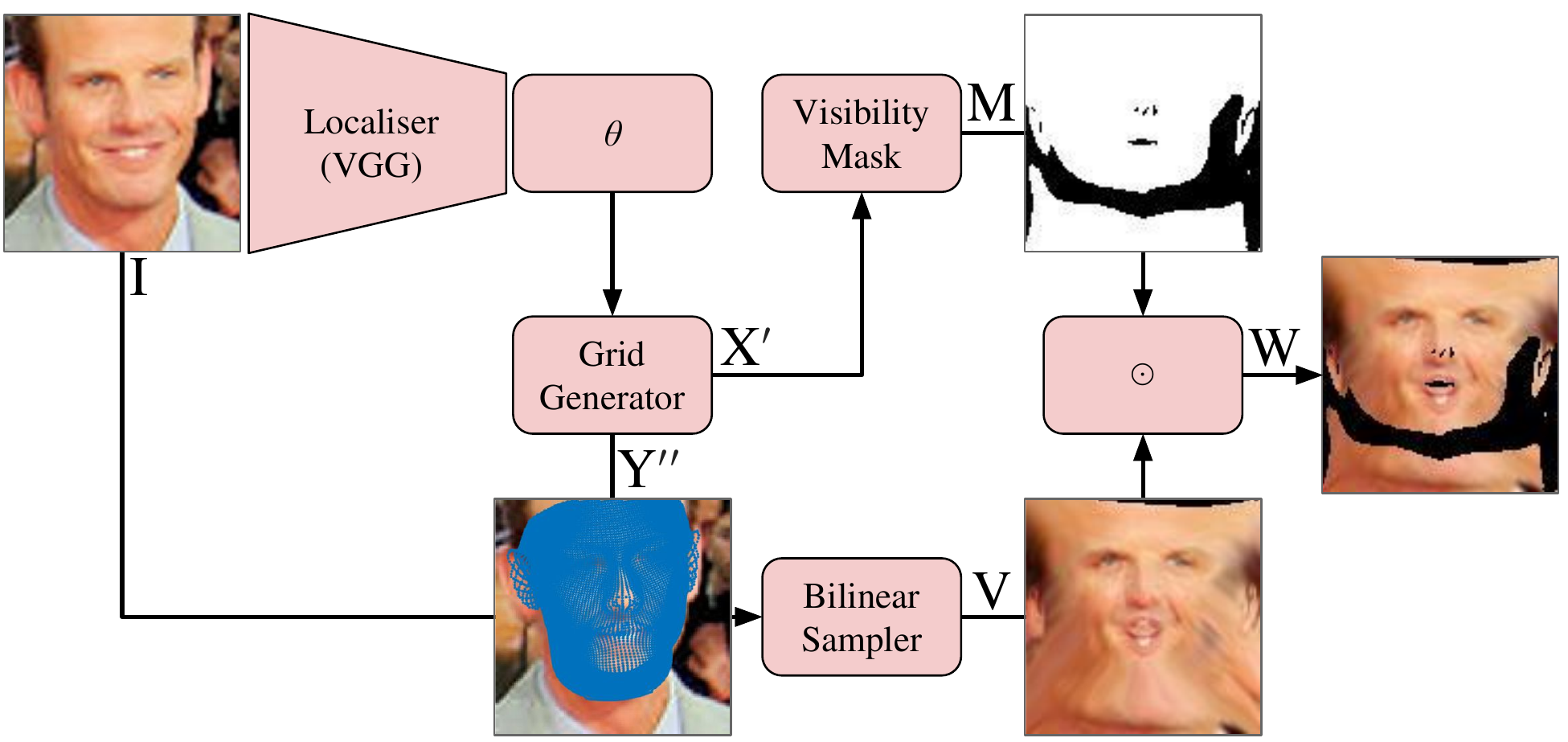}
\end{center}
\caption{Overview of the 3DMM-STN. The localiser predicts 3DMM shape parameters and pose. The grid generator projects the 3D geometry to 2D. The bilinear sampler resamples the input image to a regular output grid which is then masked by an occlusion mask computed from the estimated 3D geometry.}
\label{fig:diagram_stn}
\end{figure}

Our proposed 3DMM-STN has the same components as a conventional STN, however each component must be modified to incorporate the statistical shape model, 3D transformations and projection and self-occlusion. In this section we describe each component of a 3DMM-STN and the layers that are required to construct it. We show an overview of our architecture in Figure \ref{fig:diagram_stn}.

\subsection{Localiser network}

The localiser network is a CNN that takes an image as input and regresses the pose and shape parameters, $\theta$,  of the face in the image. Specifically, we predict the following vector of parameters:
\begin{equation}
    \theta = (\underbrace{\mathbf{r},\mathbf{t},\logs}_{\textrm{pose}},\underbrace{{\bm \alpha}}_{\textrm{shape}}).
\end{equation}
Here, $\mathbf{t}\in\R^2$ is a 2D translation, $\mathbf{r}\in\R^3$ is an axis-angle representation of a 3D rotation with rotation angle $\|\mathbf{r}\|$ and axis $\mathbf{r}/\|\mathbf{r}\|$. Since scale must be positive, we estimate log scale and later pass this through an exponentiation layer, ensuring that the estimated scale is positive. The shape parameters ${\bm \alpha}\in\R^D$ are the principal component weights used to reconstruct the shape.

For our localiser network, we use the pretrained VGG-Faces \cite{parkhi2015deep} architecture, delete the classification layer and add a new fully connected layer with $6+D$ outputs. The weights for the new layer are randomly initialised but scaled so that the elements of the axis-angle vector are in the range $[-\pi,\pi]$ for typical inputs. The whole localiser is then fine-tuned as part of the subsequent training.

\subsection{Grid generator network}

In contrast to a conventional STN, the warped sampling grid is not obtained by applying a global transformation to the regular output grid. Instead, we apply a 3D transformation and projection to a 3D mesh that comes from the morphable model. The intensities sampled from the source image are then assigned to the corresponding points in a flattened 2D grid.

\begin{figure}[t]
\begin{center}
\resizebox{\columnwidth}{!}{%
\begin{tikzpicture}[node distance=2cm]
\node (input) [node] {Input: $\theta = (\mathbf{r},\mathbf{t},\logs,{\bm \alpha})$};
\coordinate[below of=input] (belowinput);
\node (exp) [node, right of=belowinput] {$\exp$};
\node (r2R) [node, left of=belowinput] {$\mathbf{r}$ to $\mathbf{R}$};
\node (R) [node, below of=r2R] {Rotate};
\node (3DMM) [node, left of=R] {3DMM};
\node (project) [node, right of=R] {Project};
\node (scale) [node, below of=exp] {Scale};
\node (tran) [node, right of=scale] {Translate};
\coordinate[right of=tran] (output);
\draw [arrow] (input) -- node[anchor=east] {${\bf r}$} (r2R);
\draw [arrow] (input) -- node[anchor=west] {$\logs$} (exp);
\draw [arrow] (input) -| node[anchor=east] {${\bm \alpha}$} (3DMM);
\draw [arrow] (r2R) -- node[anchor=east] {${\bf R}$} (R);
\draw [arrow] (exp) -- node[anchor=east] {$s$} (scale);
\draw [arrow] (3DMM) -- node[anchor=south] {$\mathbf{X}$} (R);
\draw [arrow] (R) -- node[anchor=south] {$\mathbf{X}^{\prime}$} (project);
\draw [arrow] (project) -- node[anchor=south] {$\mathbf{Y}$} (scale);
\draw [arrow] (scale) -- node[anchor=south] {$\mathbf{Y}^{\prime}$} (tran);
\draw [arrow] (tran) -- node[anchor=south] {$\mathbf{Y}^{\prime\prime}$} (output);
\end{tikzpicture}
}
\end{center}
\caption{The grid generator network within a 3DMM-STN.}
\label{fig:gridgen}
\end{figure}
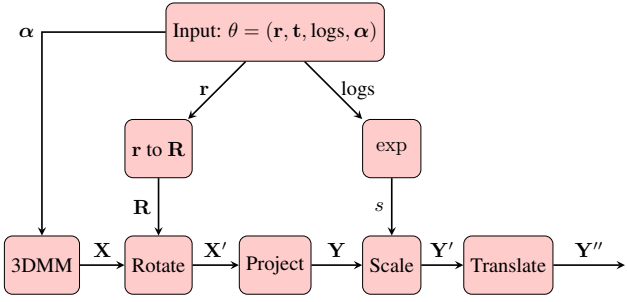

For this reason, the grid generator network in a 3DMM-STN is more complex than in a conventional STN, although we emphasise that it remains differentiable and hence suitable for use in end to end training. The sample points in our grid generator are determined by the transformation parameters $\theta$ estimated by the localiser network. Our grid generator combines a linear statistical model with a scaled orthographic projection as shown in Figure \ref{fig:gridgen}. Note that we could alternatively use a perspective projection (modifying the localiser to predict a 3D translation as well as camera parameters such as focal length). However, recent results show that interpreting face shape under perspective is ambiguous \cite{bas2017what,smith2016perspective} and so we use the more restrictive orthographic model here.

We now describe the transformation applied by each layer in the grid generator and provide derivatives.

\paragraph{3D morphable model layer}

The 3D morphable model layer generates a shape ${\bf X}\in\R^{3\times N}$ comprising $N$ 3D vertices by taking a linear combination of $D$ basis shapes (principal components) stored in the matrix $\mathbf{P}\in\R^{3N\times D}$ and the mean shape ${\bm \mu}\in\R^{3N}$ according to shape parameters ${\bm \alpha}\in\R^D$:
\begin{equation*}
    {\bf X}({\bm \alpha})_{i,j} = \mathbf{x}({\bm \alpha})_{3(j-1)+i},\quad i\in[1,3],j\in[1,N],
\end{equation*}    
where
\begin{equation*}
    \mathbf{x}({\bm \alpha}) = {\bf P}{\bm \alpha} + {\bm \mu}
\end{equation*}
and the derivatives are given by:
\begin{equation*}
    \frac{\partial\mathbf{x}}{\partial {\bm \alpha}}=\mathbf{P},\quad \frac{\partial X_{i,j}}{\partial \alpha_k}=P_{3(j-1)+i,k}.
\end{equation*}
Note that such a linear model is exactly equivalent to a fully connected layer (and hence a special case of a convolutional layer) with fixed weights and biases. This is not at all surprising since a linear model is exactly what is implemented by a single layer linear decoder. In this interpretation, the shape parameters play the role of the input map, the principal components the role of weights and the mean shape the role of biases. This means that this layer can be implemented using an existing implementation of a convolution layer and also, following our later suggestion for future work, that the model could itself be made trainable simply by having non-zero learning rate for the convolution layer. 

In our network, we use some of the principal components to represent shape variation due to identity and the remainder to represent deformation due to expression. We assume that expressions are additive and we can thus combine the two into a single linear model. Note that the shape parameters relating to identity may contain information that is useful for recognition, so these could be incorporated into a descriptor in a recognition network after the STN.

\paragraph{Axis-angle to rotation matrix layer}

This layer converts an axis-angle representation of a rotation, $\mathbf{r}\in\R^3$, into a rotation matrix:
\begin{equation*}
    \mathbf{R}(\mathbf{r}) = \cos\theta\mathbf{I} + \sin\theta \begin{bmatrix} \barr \end{bmatrix}_{\times} + (1-\cos\theta)\barr\barr^T,
\end{equation*}
where $\theta=\|\mathbf{r}\|$ and $\barr=\mathbf{r} / \| \mathbf{r} \|$ and
\begin{equation*}
\begin{bmatrix} \mathbf{a} \end{bmatrix}_{\times} = \begin{bmatrix} 0 & -a_3 & a_2 \\ a_3 & 0 & -a_1 \\ -a_2 & a_1 & 0 \end{bmatrix}
\end{equation*}
is the cross product matrix. The derivatives are given by \cite{gallego2015compact}:
\begin{equation*}
    \frac{\partial\mathbf{R}}{\partial r_i} = 
    \begin{cases}
    \left[ \mathbf{e}_i \right]_{\times} & \textrm{if}\ \mathbf{r}=\mathbf{0} \\
    \frac{r_i\left[ \mathbf{r} \right]_{\times} + \left[ \mathbf{r}\times({\bf I}-\mathbf{R}(\mathbf{r}))\mathbf{e}_i \right]_{\times}}{\|\mathbf{r}\|^2}\mathbf{R} & \textrm{otherwise}
    \end{cases}
\end{equation*}
where $\mathbf{e}_i$ is the $i$th vector of the standard basis in $\R^3$.

\paragraph{3D rotation layer}

The rotation layer takes as input a rotation matrix $\mathbf{R}$ and $N$ 3D points $\mathbf{X}\in\R^{3\times N}$ and applies the rotation:
\begin{gather*}
    \mathbf{X}^{\prime}(\mathbf{R},\mathbf{X})=\mathbf{RX}\\
    \frac{\partial X^{\prime}_{i,j}}{\partial R_{i,k}}=X_{k,j},\quad \frac{\partial X^{\prime}_{i,j}}{\partial X_{k,j}}=R_{i,k},\quad i,k\in[1,3], j\in[1,N].
\end{gather*}

\paragraph{Orthographic projection layer}

The orthographic projection layer takes as input a set of $N$ 3D points $\mathbf{X}^{\prime}\in\R^{3\times N}$ and outputs $N$ 2D points $\mathbf{Y}\in\R^{2\times N}$ by applying an orthographic projection along the z axis:
\begin{gather*}
    {\bf Y}(\mathbf{X}^{\prime})=\mathbf{PX}^{\prime},\quad \mathbf{P}=\begin{bmatrix}1&0&0\\0&1&0\end{bmatrix},\\
    \frac{\partial Y_{i,j}}{\partial X^{\prime}_{i,j}}=1,\quad i\in[1,2], j\in[1,N].
\end{gather*}

\paragraph{Scaling}

The log scale estimated by the localiser is first transformed to scale by an exponentiation layer:
\begin{equation*}
    s(\logs)=\exp(\logs),\quad \frac{\partial s}{\partial \logs}=\exp(\logs).
\end{equation*}
Then, the 2D points $\mathbf{Y}\in\R^{2\times N}$ are scaled:
\begin{equation*}
    \mathbf{Y}^{\prime}(s,\mathbf{Y})=s\mathbf{Y},\quad \frac{\partial Y^{\prime}_{i,j}}{\partial s}=Y_{i,j},\quad \frac{\partial Y^{\prime}_{i,j}}{\partial Y_{i,j}}=s
\end{equation*}

\paragraph{Translation}

Finally, the 2D sample points are generated by adding a 2D translation $\mathbf{t}\in\R^2$ to each of the scaled points:
\begin{equation*}
    \mathbf{Y}^{\prime\prime}(\mathbf{t},\mathbf{Y}^{\prime})=\mathbf{Y}^{\prime}+\mathbf{1}_{N}\otimes \mathbf{t},\quad \frac{\partial Y^{\prime\prime}_{i,j}}{\partial t_i}=1,\quad \frac{\partial Y^{\prime\prime}_{i,j}}{\partial Y^{\prime}_{i,j}}=1,
\end{equation*}
where $\mathbf{1}_{N}$ is the row vector of length $N$ containing ones and $\otimes$ is the Kronecker product.

\subsection{Sampling}

In the original STN, the sampler component used bilinear sampling to sample values from the input image and transform them to an output grid. We make a number of modifications. First, the output grid is a texture space flattening of the 3DMM mesh. Second, the bilinear sampler layer will incorrectly sample parts of the face onto vertices that are self-occluded so we introduce additional layers that calculate which vertices are occluded and mask the sampled image appropriately.

\paragraph{Output grid}
The purpose of an STN is to transform an input image into a canonical, pose-normalised view. In the context of a 3D model, one could imagine a number of analogous ways that an input image could be normalised. For example, the output of the STN could be a rendering of the mean face shape in a frontal pose with the sampled texture on the mesh. Instead, we choose to output sampled textures in a 2D embedding obtained by flattening the mean shape of the 3DMM. This ensures that the output image is approximately area uniform with respect to the mean shape and also that the whole output image contains face information.

\begin{figure}[t]
\begin{center}
\includegraphics[clip, height=2.5cm]{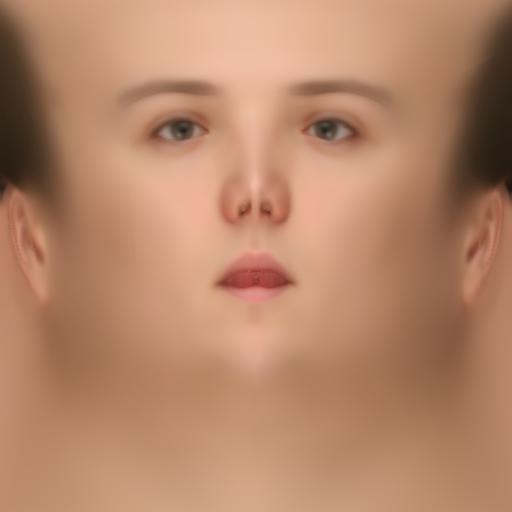}
\includegraphics[clip, height=2.5cm]{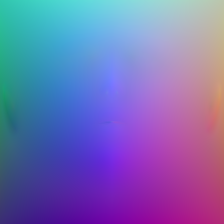}
\end{center}
\caption{The output grid of our 3DMM-STN: a Tutte embedding of the mean shape of the Basel Face Model. On the left we show a visualisation using the mean texture (though note that our 3DMM-STN does not use a texture model). On the right we show the mean shape as a geometry image \cite{gu2002geometry}.
}
\label{fig:UVspace}
\end{figure}

Specifically, we compute a Tutte embedding \cite{floater1997parametrization} using conformal Laplacian weights and with the mesh boundary mapped to a square. To ensure a symmetric embedding we map the symmetry line to the symmetry line of the square, flatten only one side of the mesh and obtain the flattening of the other half by reflection. We show a visualisation of our embedding using the mean texture in Figure \ref{fig:UVspace}. In order that the output warped image produces a regularly sampled image, we regularly re-sample (i.e. re-mesh) the 3DMM (mean and principal components) over a uniform grid of size $H^{\prime}\times W^{\prime}$ in this flattened space. This effectively makes our 3DMM a deformable geometry image \cite{gu2002geometry}. The re-sampled 3DMM that we use in our STN therefore has $N=H^{\prime}W^{\prime}$ vertices and each vertex $i$ has an associated UV coordinate $(x_i^t,y_i^t)$. The corresponding sample coordinate produced by the grid generator is given by $(x_i^s,y_i^s)=(Y^{\prime\prime}_{1,i},Y^{\prime\prime}_{2,i})$. 

\paragraph{Bilinear sampling}
We use bilinear sampling, exactly as in the original STN such that the re-sampled image $V_i^c$ at location $(x_i^t,y_i^t)$ in colour channel $c$ is given by:
\begin{equation*}
    V_i^c = \sum_{j=1}^H\sum_{k=1}^W I_{jk}^c \max(0,1-|x_i^s-k|)\max(0,1-|y_i^s-j|)
\end{equation*}
where $I_{jk}^c$ is the value in the input image at pixel $(j,k)$ in colour channel $c$. $I$ has height $H$ and width $W$. This bilinear sampling is differentiable (see \cite{DBLP:conf/nips/JaderbergSZK15} for derivatives) and so the loss can be backpropagated through the sampler and back into the grid generator.

\paragraph{Self-occlusions} Since the 3DMM produces a 3D mesh, parts of the mesh may be self-occluded. The occluded vertices can be computed exactly using ray-tracing or z-buffering or they can be precomputed and stored in a lookup table. For efficiency, we approximate occlusion by only computing which vertices have backward facing normals. 
This approximation would be exact for any object that is globally convex. For objects with concavities, the approximation will underestimate the set of occluded vertices. Faces are typically concave around the eyes, the nose boundary and the mouth interior but we find that typically only around 5\% of vertices are mislabelled and the accuracy is sufficient for our purposes.

This layer takes as input the rotation matrix ${\bf R}$ and the shape parameters ${\bm \alpha}$ and outputs a binary occlusion mask ${\bf M}\in\{0,1\}^{H^{\prime}\times W^{\prime}}$. The occlusion function is binary and hence not differentiable at points where the visibility of a vertex changes, everywhere else the gradient is zero. Hence, we simply pass back zero gradients:
\begin{equation*}
\frac{\partial{\bf M}}{\partial {\bm \alpha}} = 0,\quad \frac{\partial{\bf M}}{\partial {\bf R}} = 0.
\end{equation*}
Note that this means that the network is not able to learn how changes in occlusion help to reduce the loss. Occlusions are applied in a forward pass but changes in occlusion do not backpropagate.

\paragraph{Masking layer}
The final layer in the sampler combines the sampled image and the visibility map via pixel-wise products:
\begin{equation*}
W_i^c = V_i^c M_{x_i^t,y_i^t},\quad \frac{\partial W_i^c}{\partial V_i^c}=M_{x_i^t,y_i^t},\quad \frac{\partial W_i^c}{\partial M_{x_i^t,y_i^t}}=V_i^c.
\end{equation*}

\section{Geometric losses for localiser training}\label{sec:losses}

An STN is usually inserted into a network as a preprocessor of input images and its output is then passed to a classification or regression CNN. Hence, the pose normalisation that is learnt by the STN is the one that produces optimal performance on the subsequent task. In the context of a 3D morphable face model, an obvious task would be face recognition. While this is certainly worth pursuing, we have observed that the optimal normalisation for recognition may not correspond to the correct model-image correspondence one would expect. For example, if context provided by hair and clothing helps with recognition, then the 3DMM-STN may learn to sample this.

Instead, we show that it is possible to train an STN to perform accurate localisation using only some simple geometric priors without even requiring identity labels for the images. We describe these geometric loss functions in the following sections.

\begin{figure}[t]
\begin{center}
\includegraphics[clip, trim=6cm 7cm 6cm 7cm, width=1.00\columnwidth]{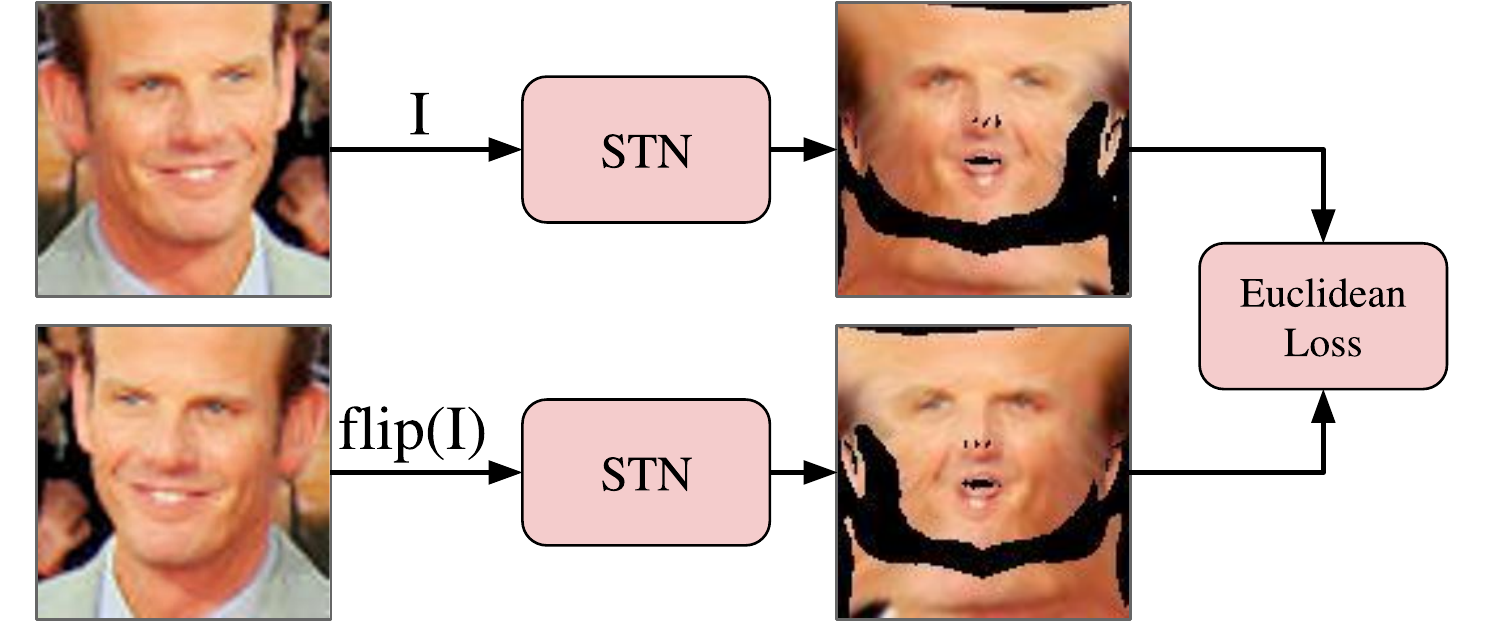}
\end{center}
\vspace*{-1mm}
\caption{Siamese multiview loss. An image and its horizontal reflection yield two sampled images. We penalise differences in these two images.}
\label{fig:diagram_siamese}
\end{figure}

\begin{figure}[b]
\begin{center}
\includegraphics[clip, trim=8cm 8.2cm 8cm 8.2cm, width=1.00\columnwidth]{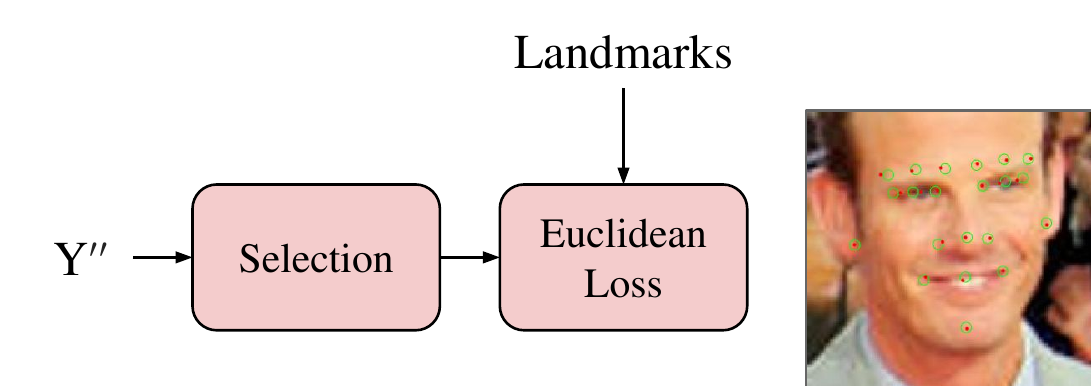}
\end{center}
\vspace*{-1mm}
\caption{Landmark loss. Left: The diagram shows the implementation of the regression layer that computes the Euclidean distance between selected 2D points and ground truth positions. Right: Predicted positions are in red and landmark positions are in green.}
\label{fig:diagram_landmark}
\end{figure}

\subsection{Bilateral symmetry loss}\label{sec:bilateral}

Faces are approximately bilaterally symmetric. Ignoring the effects of illumination, this means that we expect sampled face textures to be approximately bilaterally symmetric. We can define a loss that measures asymmetry of the sampled texture over visible pixels:
\begin{equation}
    \ell_{\textrm{sym}} = \sum_{i=1}^N \sum_{c=1}^3 M_{x_i^t,y_i^t}M_{x_{\textrm{sym}(i)}^t,y_{\textrm{sym}(i)}^t}(V_i^c - V_{\textrm{sym}(i)}^c)^2,
\end{equation}
where $V_{\textrm{sym}(i)}^c$ is the value in the resampled image at location $(W^{\prime}+1-x_i^s,y_i^s)$.

\subsection{Siamese multi-view fitting loss}\label{sec:siamese}

If we have multiple images of the same face in different poses (or equivalently from different viewpoints), then we expect that the sampled textures will be equal (again, neglecting the effects of illumination). If we had such multiview images, this would allow us to perform Siamese training where a pair of images in different poses were sampled into images $V_i^c$ and $W_i^c$ with visibility masks ${\bf M}$ and ${\bf N}$ giving a loss:
\begin{equation}
    \ell_{\textrm{multiview}} = \sum_{i=1}^N \sum_{c=1}^3 M_{x_i^t,y_i^t}N_{x_i^t,y_i^t}(V_i^c - W_i^c)^2.
\end{equation}
Ideally, this loss would be used with a multiview face database or even a face recognition database where images of the same person in different in-the-wild conditions are present. We use an even simpler variant which does not require multiview images; again based on the bilateral symmetry assumption. A horizontal reflection of a face image approximates what that face would look like in a reflected pose. Hence, we perform Siamese training on an input image and its horizontal reflection. This is different to the bilateral symmetry loss and is effectively encouraging the localiser to behave symmetrically.

\newcommand{\frontsize}{1.6cm}
\begin{figure}[!t]
\centering
\begin{tabular}{@{\hspace{0.05cm}}c@{\hspace{0.05cm}}c@{\hspace{0.05cm}}c@{\hspace{0.05cm}}c@{\hspace{0.05cm}}c@{\hspace{0.05cm}}}
\includegraphics[height=\frontsize, clip=true]{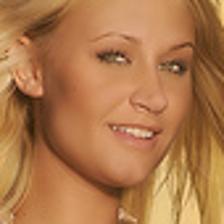}&
\includegraphics[height=\frontsize, clip=true]{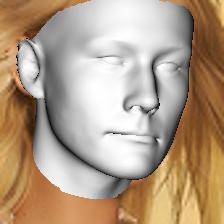}&
\includegraphics[height=\frontsize, clip=true]{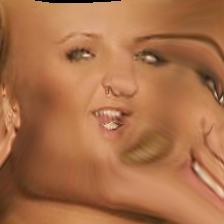}&
\includegraphics[height=\frontsize, clip=true]{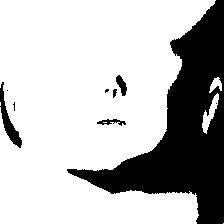}&
\includegraphics[height=\frontsize, clip=true]{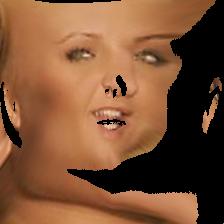}\\
\end{tabular}
\caption{Overview of the 3DMM-STN. From left to right: input image; rendering of estimated shape in estimated pose; sampled image; occlusion mask; final output of 3DMM-STN.}
\label{fig:overview}
\end{figure}

\subsection{Landmark loss}

As has been observed elsewhere \cite{tewari2017mofa}, convergence of the training can be speeded up by introducing surrogate loss functions that provide supervision in the form of landmark locations. It is straightforward to add a landmark loss to our network. First, we define a selection layer that selects $L<N$ landmarks from the $N$ 2D points outputted by the grid generator:
\begin{equation}
    \mathbf{L} = \mathbf{Y}^{\prime\prime}\mathbf{S}
\end{equation}
where $\mathbf{S}\in\{0,1\}^{N\times L}$ is a selection matrix with $\mathbf{S}^T\mathbf{S}=\mathbf{I}_L$. Given $L$ landmark locations ${\bf l}_1,\dots,{\bf l}_L$ and associated detection confidence values $c_1,\dots,c_L$, we computed a weighted Euclidean loss:
\begin{equation}
    \ell_{\textrm{landmark}} = \sum_{i=1}^L c_i\|{\bf L}_i-{\bf l}_i\|^2.
\end{equation}
Landmarks that are not visible (i.e. were not hand-labelled or detected) are simply assigned zero confidence.

\subsection{Statistical prior loss}

The statistical shape model provides a prior. We scale the shape basis vectors such that the shape parameters follow a standard multivariate normal distribution: ${\bm \alpha}\sim{\cal N}({\bf 0},{\bf I}_D)$. Hence, the statistical prior can be encoded by the following loss function:
\begin{equation}
    \ell_{\textrm{prior}} = \|{\bm \alpha}\|^2.
\end{equation}

\section{Experiments}

\begin{figure}[!t]
\centering
\begin{tabular}{@{\hspace{0.05cm}}c@{\hspace{0.05cm}}c@{\hspace{0.05cm}}c@{\hspace{0.05cm}}c@{\hspace{0.05cm}}c@{\hspace{0.05cm}}}
\includegraphics[height=\frontsize, clip=true]{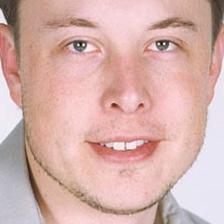}&
\includegraphics[height=\frontsize, clip=true]{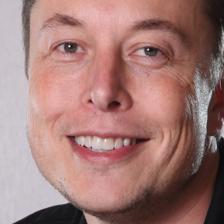}&
\includegraphics[height=\frontsize, clip=true]{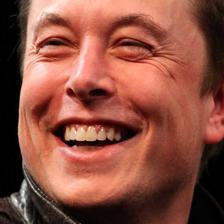}&
\includegraphics[height=\frontsize, clip=true]{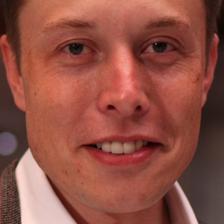}&
\includegraphics[height=\frontsize, clip=true]{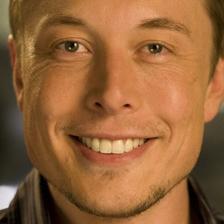}\\
\includegraphics[height=\frontsize, clip=true]{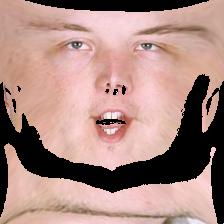}&
\includegraphics[height=\frontsize, clip=true]{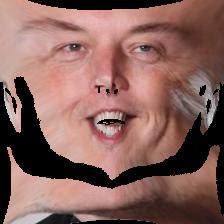}&
\includegraphics[height=\frontsize, clip=true]{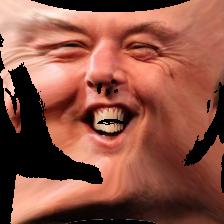}&
\includegraphics[height=\frontsize, clip=true]{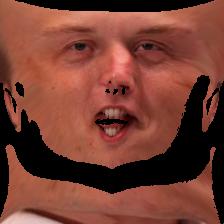}&
\includegraphics[height=\frontsize, clip=true]{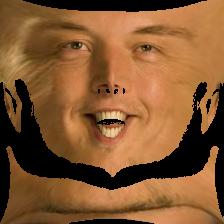}\\
\end{tabular}
\caption{3DMM-STN output for multiple images of the same person in different poses.}
\label{fig:qualitative_grid}
\end{figure}

For our statistical shape model, we use $D=10$ dimensions of which five are the first five (identity) principal components from the Basel Face Model \cite{DBLP:conf/avss/PaysanKARV09}. The other five are expression components which come from FaceWarehouse \cite{cao2014facewarehouse} using the correspondence to the Basel Model provided by \cite{zhu2016face}. We re-mesh the Basel Model over a uniform grid of size $224\times 224$. We trained our 3DMM-STN with the four loss functions described in Section \ref{sec:losses} using the AFLW database \cite{koestinger11a}. This provides up to 21 landmarks per subject for over 25k in-the-wild images. This is a relatively small dataset for training a deep network so we perform `fine-tuning' by setting the learning rate on the last layer of the localiser to four times that of the rest of the network. 
Figure \ref{fig:overview} shows the pipeline of an image passing through a 3DMM-STN.
A by-product of the trained 3DMM-STN is that it can also act as a 2D landmark localiser. After training, the localiser achieves an average landmarking error of 2.35 pixels on the part of AFLW used as validation set, over the 21 landmarks, showing that overall, the training converges well.

We begin by demonstrating that our 3DMM-STN learns to predict consistent correspondence between model and image. In Figure \ref{fig:qualitative_grid} we show 3DMM-STN output for multiple images of the same person. Note that the features are consistently mapped to the same location in the transformed output. In Figure \ref{fig:meanfaces} we go further by applying the 3DMM-STN to multiple images of the same person and then average the resulting transformed images. We show results for 10 subjects from the UMDFaces \cite{bansal2016umdfaces} dataset. The number of images for each subject is shown in parentheses. The averages have well-defined features despite being computed from images with large pose variation.

In Figure \ref{fig:Compflatim} we provide a qualitative comparison to \cite{tran2017regressing}. This is the only previous work on 3DMM fitting using a CNN for which the trained network is made publicly available. In columns one and five, we show input images from UMDFaces \cite{bansal2016umdfaces}. In columns two and six, we show the reconstruction provided by \cite{tran2017regressing}. While the reconstruction captures the rough appearance of the input face, it lacks the discriminating detail of the original image. This method regresses shape and texture directly but not illumination or pose. Hence, we cannot directly compare the model-image correspondence provided by this method. To overcome this, we use the landmark detector used by \cite{tran2017regressing} during training and compute the optimal pose to align their reconstruction to these landmarks. We replace their cropped model by the original BFM shape model and sample the image. This allows us to create the flattened images in columns three and seven. The output of our proposed 3DMM-STN is shown in columns four and eight. We note that our approach less frequently samples background and yields a more consistent correspondence of the resampled faces. In the bottom row of the figure we show challenging examples where \cite{tran2017regressing} did not produce any output because the landmark detector failed. Despite occlusions and large out of plane rotations, the 3DMM-STN still does a good job of producing a normalised output image.

\newcommand{\averagesize}{1.6cm}
\begin{figure}[!t]
\centering
\begin{tabular}{@{\hspace{0.03cm}}c@{\hspace{0.03cm}}c@{\hspace{0.03cm}}c@{\hspace{0.03cm}}c@{\hspace{0.03cm}}c@{\hspace{0.03cm}}}

\includegraphics[height=\averagesize, clip=true]{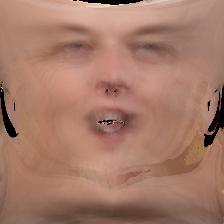}&
\includegraphics[height=\averagesize, clip=true]{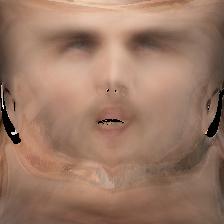}&
\includegraphics[height=\averagesize, clip=true]{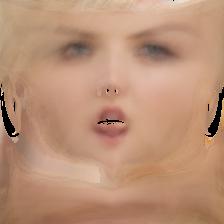}&
\includegraphics[height=\averagesize, clip=true]{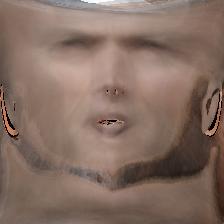}&
\includegraphics[height=\averagesize, clip=true]{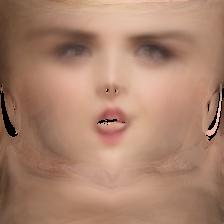}\\[-0.5em]
\tiny{Elon Musk (34)}&\tiny{Christian Bale (51)}&\tiny{Elisha Cuthbert (53)}&\tiny{Clint Eastwood (62)}&\tiny{Emma Watson (73)}\\
\includegraphics[height=\averagesize, clip=true]{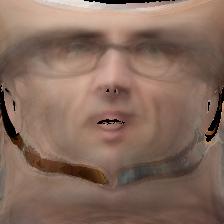}&
\includegraphics[height=\averagesize, clip=true]{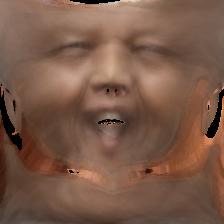}&
\includegraphics[height=\averagesize, clip=true]{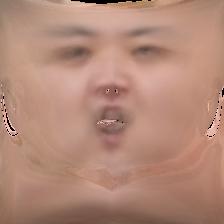}&
\includegraphics[height=\averagesize, clip=true]{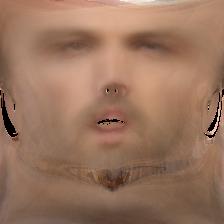}&
\includegraphics[height=\averagesize, clip=true]{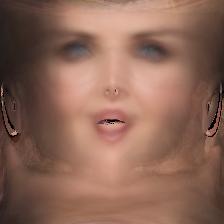}\\[-0.5em]

\tiny{Chuck Palahniuk (48)}& \tiny{Nelson Mandela (52)}&\tiny{Kim Jong-un (60)}&\tiny{Ben Affleck (66)}&\tiny{Courteney Cox (127)} \\
\end{tabular}
\caption{A set of mean flattened images per subject. Real images are obtained from UMDFaces dataset. The number of images that are used for averaging is stated next to subject's name.}
\label{fig:meanfaces}
\end{figure}

\newcommand{\frontimsize}{2.1cm}
\newcommand{\meshsize}{1.6cm}
\setlength{\tabcolsep}{1pt}
\begin{figure*}[!t]
{\small
\centering
\begin{tabular}{cccc@{\hspace{0.1cm}}|@{\hspace{0.1cm}}cccc}
Input & \cite{tran2017regressing} & \cite{tran2017regressing} Flatten & 3DMM-STN & Input & \cite{tran2017regressing} & \cite{tran2017regressing} Flatten & 3DMM-STN \\
\includegraphics[height=\frontimsize, clip=true]{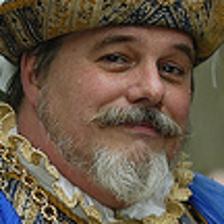}&
\raisebox{0.2cm}{\includegraphics[height=\meshsize, clip=true, trim=130px 60px 100px 30px]{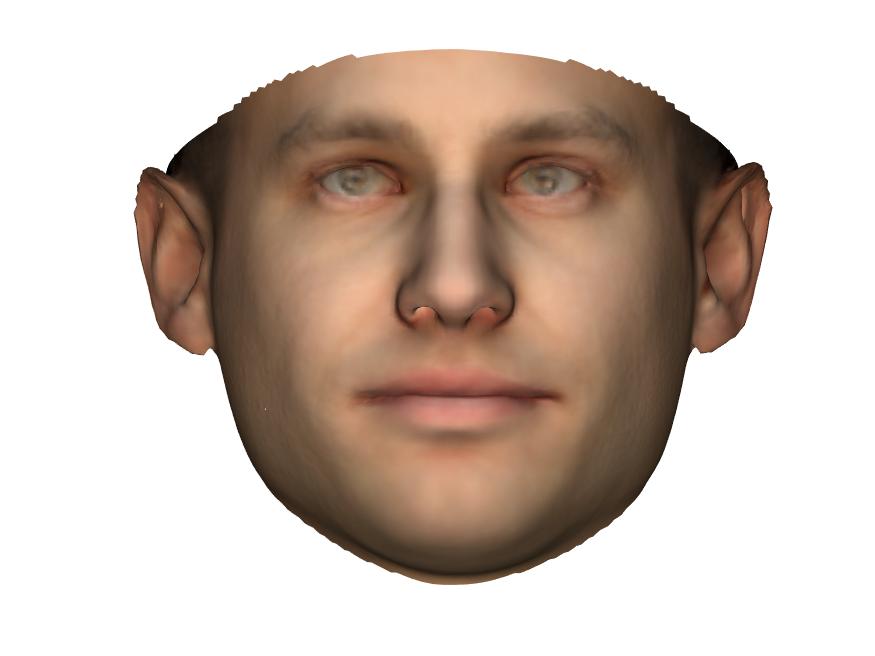}}&
\includegraphics[height=\frontimsize, clip=true]{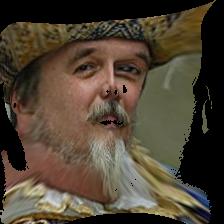}&
\includegraphics[height=\frontimsize, clip=true]{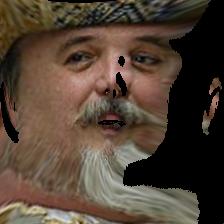}
&
\includegraphics[height=\frontimsize, clip=true]{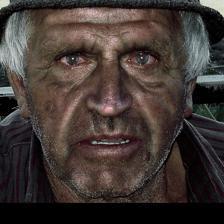}&
\raisebox{0.2cm}{\includegraphics[height=\meshsize, clip=true, trim=130px 60px 100px 30px]{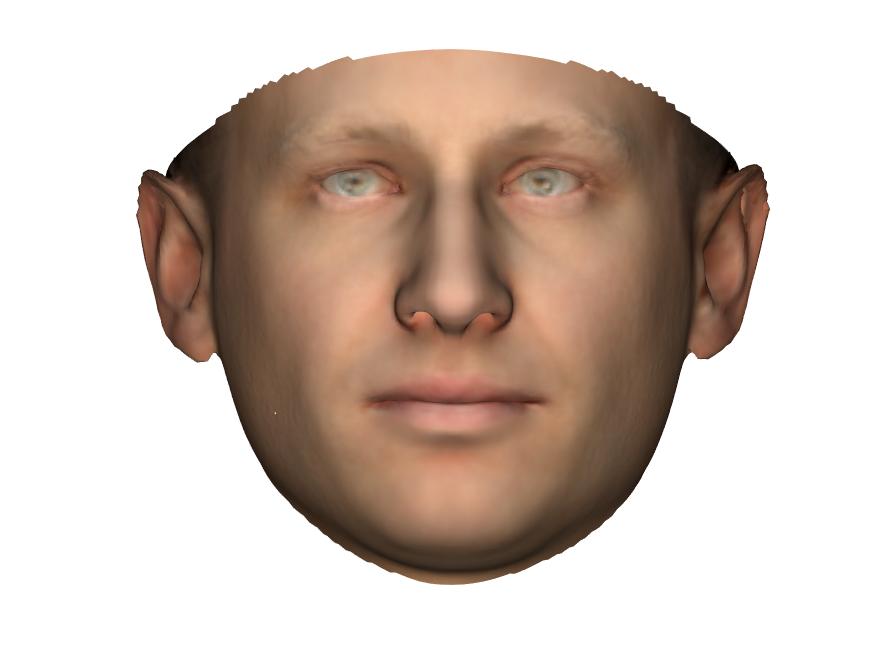}}&
\includegraphics[height=\frontimsize, clip=true]{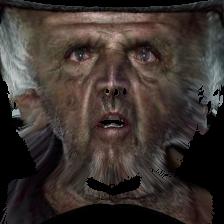}&
\includegraphics[height=\frontimsize, clip=true]{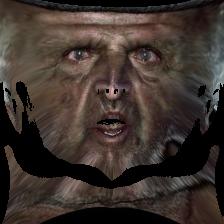}
\\
\includegraphics[height=\frontimsize, clip=true]{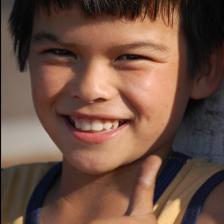}&
\raisebox{0.2cm}{\includegraphics[height=\meshsize, clip=true, trim=100px 60px 75px 30px]{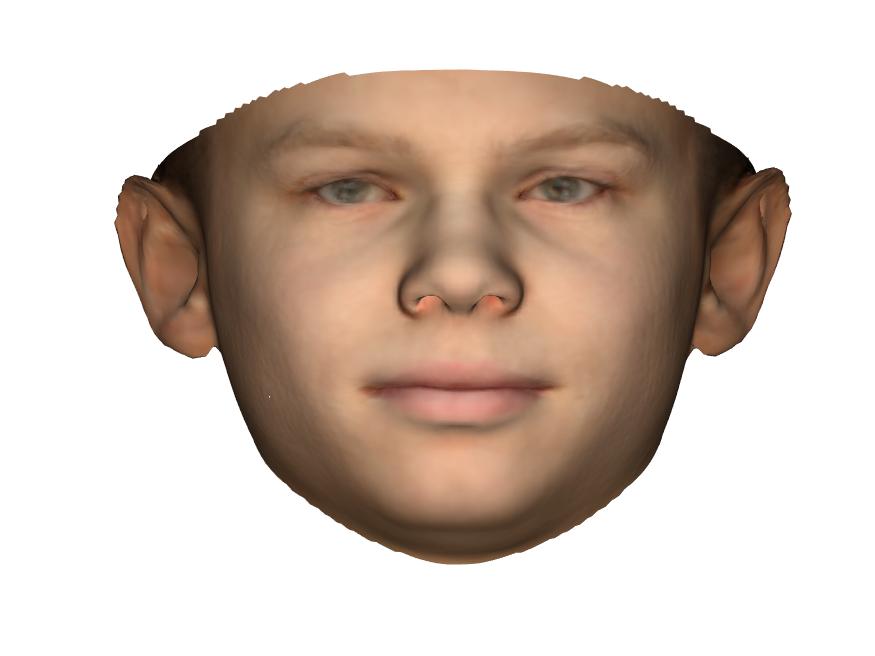}}&
\includegraphics[height=\frontimsize, clip=true]{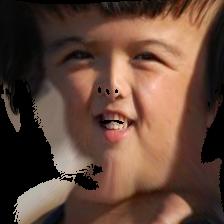}&
\includegraphics[height=\frontimsize, clip=true]{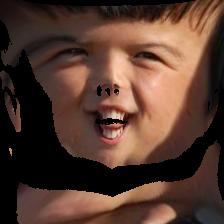}
&
\includegraphics[height=\frontimsize, clip=true]{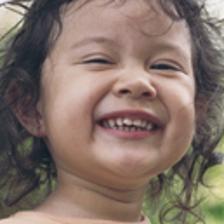}&
\raisebox{0.2cm}{\includegraphics[height=\meshsize, clip=true, trim=130px 60px 100px 30px]{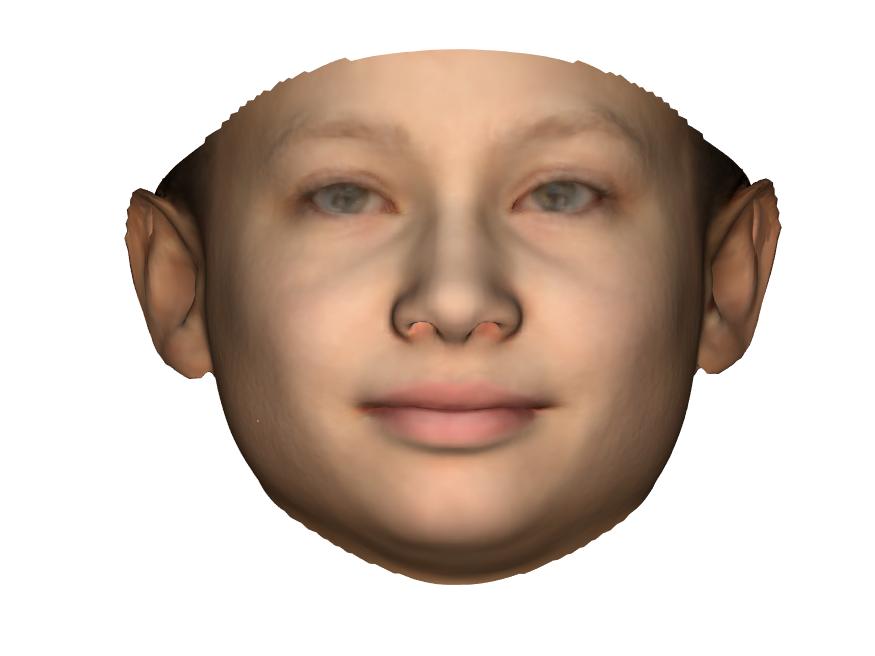}}&
\includegraphics[height=\frontimsize, clip=true]{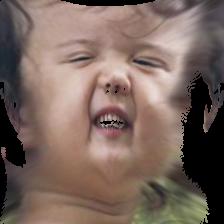}&
\includegraphics[height=\frontimsize, clip=true]{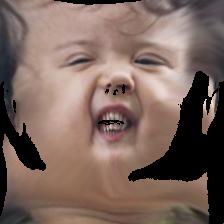}
\\
\includegraphics[height=\frontimsize, clip=true]{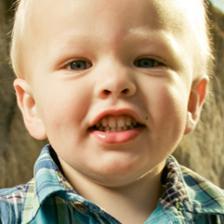}&
\raisebox{0.2cm}{\includegraphics[height=\meshsize, clip=true, trim=130px 60px 100px 30px]{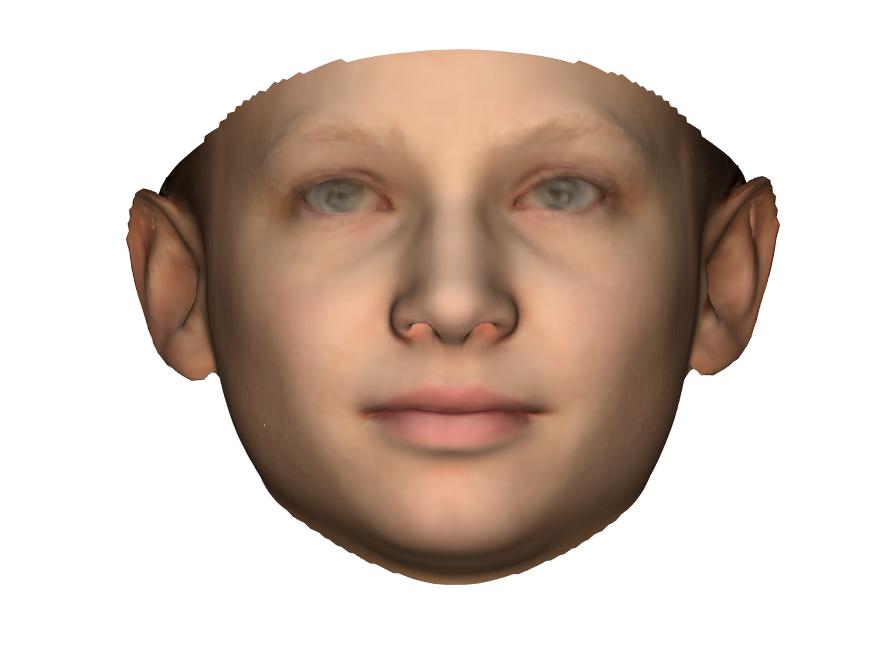}}&
\includegraphics[height=\frontimsize, clip=true]{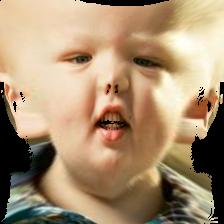}&
\includegraphics[height=\frontimsize, clip=true]{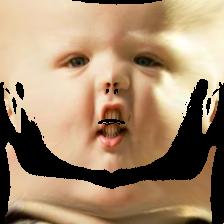}
&
\includegraphics[height=\frontimsize, clip=true]{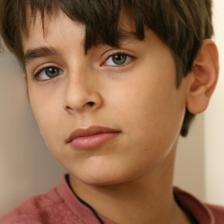}&
\raisebox{0.2cm}{\includegraphics[height=\meshsize, clip=true, trim=100px 60px 75px 30px]{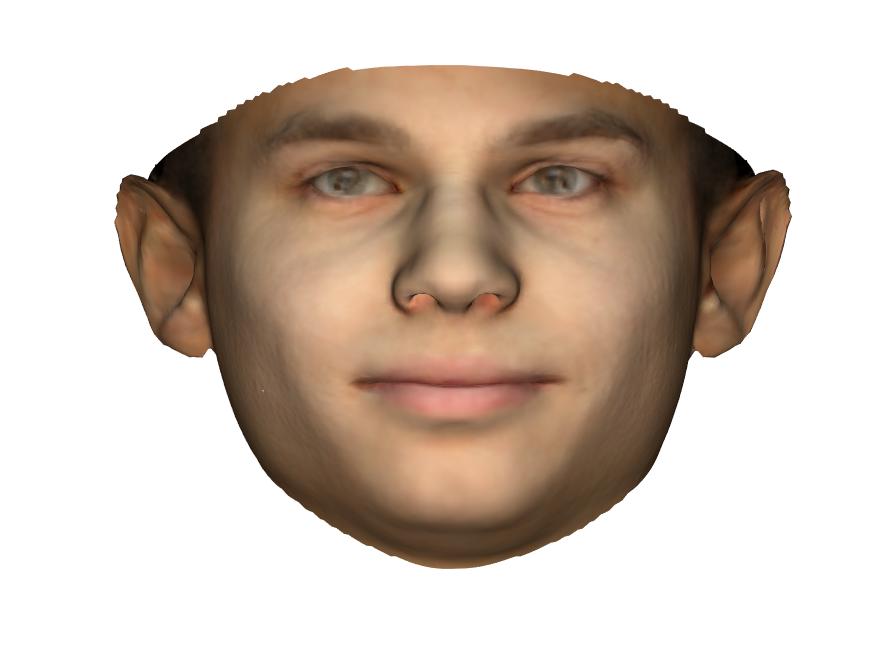}}&
\includegraphics[height=\frontimsize, clip=true]{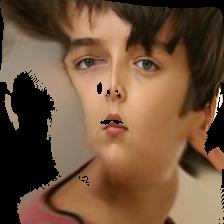}&
\includegraphics[height=\frontimsize, clip=true]{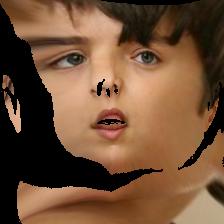}
\\
\includegraphics[height=\frontimsize, clip=true]{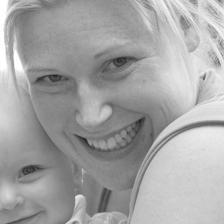}&
\raisebox{0.2cm}{\includegraphics[height=\meshsize, clip=true, trim=130px 60px 100px 30px]{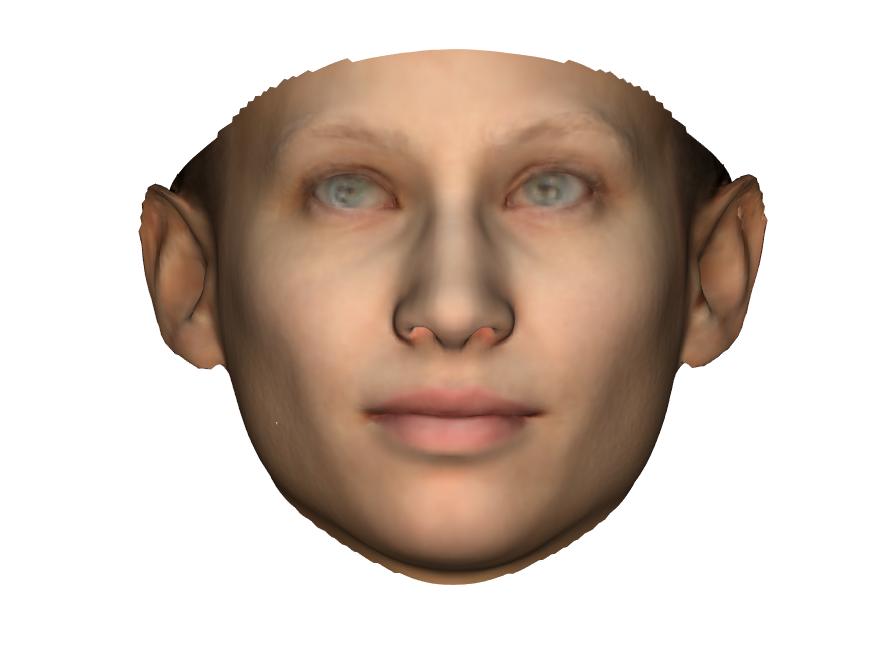}}&
\includegraphics[height=\frontimsize, clip=true]{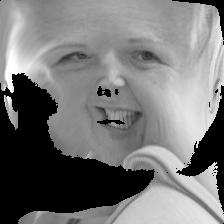}&
\includegraphics[height=\frontimsize, clip=true]{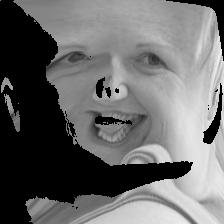}
&
\includegraphics[height=\frontimsize, clip=true]{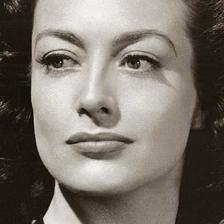}&
\raisebox{0.2cm}{\includegraphics[height=\meshsize, clip=true, trim=130px 60px 100px 30px]{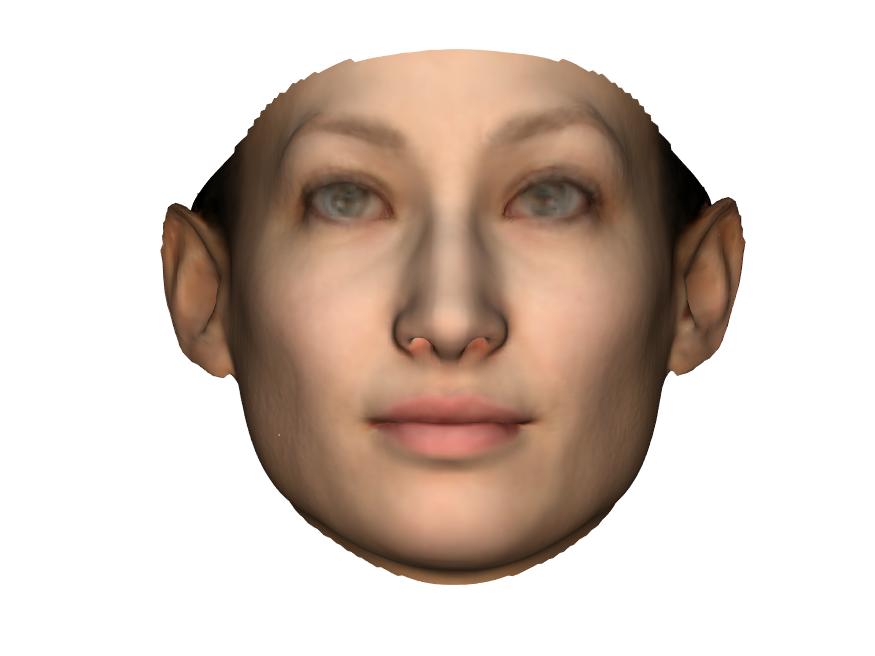}}&
\includegraphics[height=\frontimsize, clip=true]{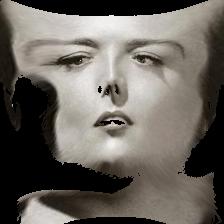}&
\includegraphics[height=\frontimsize, clip=true]{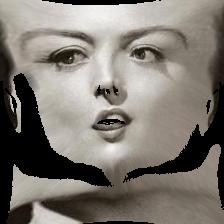}
\\\hline
Input & 3DMM-STN & Input & 3DMM-STN & Input & 3DMM-STN & Input & 3DMM-STN \\
\includegraphics[height=\frontimsize, clip=true]{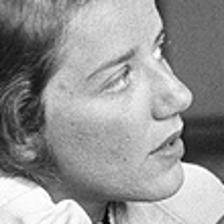}&
\includegraphics[height=\frontimsize, clip=true]{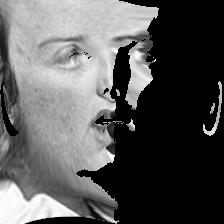}&
\includegraphics[height=\frontimsize, clip=true]{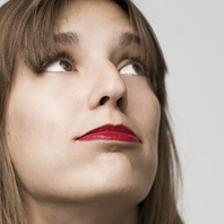}&
\includegraphics[height=\frontimsize, clip=true]{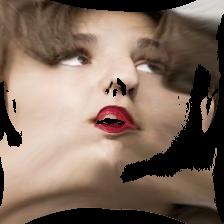}
&
\includegraphics[height=\frontimsize, clip=true]{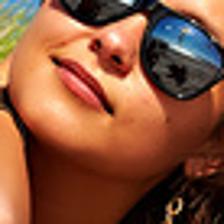}&
\includegraphics[height=\frontimsize, clip=true]{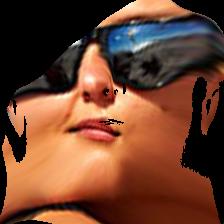}&
\includegraphics[height=\frontimsize, clip=true]{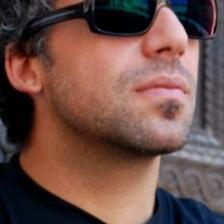}&
\includegraphics[height=\frontimsize, clip=true]{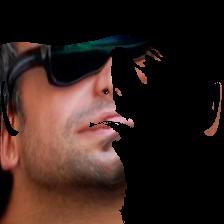}
\end{tabular}}
\caption{Qualitative comparison to \cite{tran2017regressing}. The bottom row shows examples for which \cite{tran2017regressing} failed to fit due to failure of the landmark detector.}
\label{fig:Compflatim}
\end{figure*}

\section{Conclusions}

In this paper we have shown how to use a 3D morphable model as a spatial transformer within a CNN. Our proposed architecture has a number of interesting properties. First, the network (specifically, the localiser part of the network) learns to fit a 3D morphable model to a single 2D image without needing labelled examples of fitted models. Since the problem of fitting a morphable model to an image is an unsolved problem (and therefore no existing algorithm could be assumed to provide reliable ground truth fits), this kind of unsupervised learning is desirable. 
Second,
the morphable model itself is fixed in our current architecture. However, there is no reason that this could not also be learnt. In this way, it may be possible to learn a 3D deformable model for an object class simply from a collection of images that are labelled appropriately for the chosen proxy task.

There are many ways that this work can be extended. First, we would like to investigate training our 3DMM-STN in an end to end recognition network. We would hope that the normalisation means that a recognition network could be trained on less data and with less complexity than existing networks that must learn pose invariance implicitly. Second, the shape parameters estimated by the localiser may contain discriminative information and so these could be combined into subsequent descriptors for recognition. Third, we would like to further explore the multiview fitting loss. Using a multiview face database or video would provide a rich source of data for learning accurate localisation. Finally, the possibility of learning the shape model during training is exciting and we would like to explore other objects classes besides faces for which 3DMMs do not currently exist.



\section*{Acknowledgements}

We gratefully acknowledge the support of NVIDIA Corporation with the donation of the Titan X Pascal GPU used for this research.


{\small
\bibliographystyle{ieee}
\bibliography{GMDL}
}

\end{document}